# Could robots be regarded as humans in future?

**Authors:** Huansheng Ning[1]*, Feifei Shi[1].

**Affiliations:**

[1]University of Science and Technology Beijing, Beijing, 100083, China.

*Correspondence to: ninghuansheng@ustb.edu.cn.

Abstract:

With the overwhelming advances in Artificial Intelligence (AI), brain science and neuroscience, robots are developing towards a direction of much more human-like and human-friendly. We can't help but wonder whether robots could be regarded as humans in future? In this article, we propose a novel perspective to analyze the essential difference between humans and robots, that is based on their respective living spaces, particularly the independent and intrinsic thinking space. We finally come to the conclusion that, only when robots own the independent and intrinsic thinking space as humans, could they have the prerequisites to be regarded as humans.

So far, robots are more human-like in biometric appearances, cognitive abilities, and in more cases, they are poised to be much more perceptive and dexterous, so as to handle increasingly meaningful and complicated tasks. As reported by the International Federation of Robotics, almost 2.7 billion robots are working alongside humans in factories around the global, that means the robots are being deeply trained with necessary skills and professions, and to a large extent, are fully capable of displacing countless humans with laborious jobs [1]. However, in addition to the unimagined progress and opportunities, the robot's revolution also engenders increasing challenges and threats, which has led to a lot of concerns and discussions. What happens if robots become as intelligent as humans? Could robots be regarded as humans in future?

Engineers, scientists, and anthropologists have been working hard to explore the difference between humans and robots, in particular the advanced humanoid robots in future. At this stage, most people regard self-consciousness as the main difference between humans and robots [2]. Does it mean that if one day the robot owns self-consciousness, it can be regarded as a human being? If not, what is the essential difference between robots and humans? Upon this, we continue a further discussion in this work.

## The ability of evolution or self-learning

Firstly, the ability of evolution or self-learning is reckoned as one of the differences between humans and robots. It is generated from the initial stage where robots could only handle simple tasks or missions as predefined, such as holding a bottle, catching given goods etc. However, benefiting from the overwhelming developments of Artificial Intelligence (AI) and machine learning, robots nowadays begin to own the ability of self-learning or evolution, in which they could learn unknown knowledge from large unlabeled datasets, and are able to adapt to dynamic and unpredicted circumstances autonomously [3].The gap in self-learning capabilities between humans and robots seems to be narrowing, or even closing.

# Biometric features and self-consciousness

Following that, biometric features and self-consciousness are also distinctive characteristics between humans and robots. Regarding biometric features, robots are moving in a direction that is more human-like and human-friendly, for example, soft robotics is an emerging subfield which is inspired by nature and living organisms. Instead of hard, sharp, non-deformable materials, soft robots are adopting highly compliant ones, so as to be much more adaptable to complicated tasks, as well as much safer when working with humans. Cutting-edge research on robots' biometric features are always on the way, for example, Shih overviews the state-of-the-art electronic skins in soft robots, and aims to build more autonomous, deployable, and adaptive robots combining with machine learning [4]. Rafsanjani exploits the usage of metamaterials in soft robots, which has been demonstrated with high efficiency in improving the performance [5]. In other words, with advances in materials and biological science, the skin, appearance, organs and other biological characteristics of robots are approaching human beings' step by step. It is not easy to distinguish robots from humans only depending on biometric features any more.

Another well-known and significant difference between humans and robots is self-consciousness. Generally speaking, it is a common sense that humans are superior at intuition, cognition and awareness due to its inherent brain structures and mechanisms. In recent years, with the technological breakthroughs in brain science and neuroscience, scientists also concentrate on the self-consciousness of robots. For example, Takeno has begun the related research since 2005, by demonstrating that robots could recognize themselves in the mirror as what humans do [6]. It is worth noting that for robots, the rudimentary consciousness and awareness are still in infancy, and more advanced study is being push forward steadily. As Lipson said, achieving self-simulate of robots is the first step forward self-consciousness, and we believe that with the advent of robot autonomy, self-consciousness is entirely possible in future [7].

**To be honest, due to the unprecedented developments in AI, materials science and neuroscience, the differences between humans and robots in aspects of the ability of self-learning, biometric features, and self-consciousness are gradually narrowing. Does that mean that robots could be regarded as humans one day? The answer is NO, at least for now, since we regard humans and robots have essential difference in view of their living spaces.**

# The essential difference between humans and robots in view of their living spaces

From the aspect of basic living spaces, humans are regarded to live in a Cyber-Physical-Social-Thinking Hyperspace [8], evolving from the earliest existence-based physical space, relationship-inspired social space, to the brain-abstracted thinking space, and intelligence-enabled cyberspace. However, for robots, they mainly live within Cyber-Physical-Social space, since there is no independent and explicit thinking space, with their "brains" totally being attached to cyberspace.

Compared with robots, the independent and intrinsic thinking space are exclusive characteristics for humans, and cannot be achieved by robots at present. Although intelligent techniques could allow robots imitating human intelligence to a large extent, it is still far away for robots to own such an independent thinking space as humans. That is to say, for humans, they own the totally independent thinking space, with no necessity to establish connections with outside, and can carry out brain activities themselves, while for robots, their "thoughts", "ideas" and other kind of activities, must be connected to cyberspace, with no explicit, independent, and comprehensive thinking space. In addition, the intrinsic nature of human thinking space enables thinking space to be born with, while for robots, the thinking space must be designed and implanted by external manufacturers. Both independent and intrinsic characteristics allow human thinking space to be strong, comprehensive, robust and autonomous enough for generating ideas, carrying out complicated tasks, and adapting to dynamic situations. Only by overcoming the two essential yet basic characteristics could robots have the prerequisites to be regarded as humans.

To guide and inspire more researches, Table 1 lists occasions when a robot could be regarded as human or evolved into a truly human being. If the subject has the ability of evolution or self-learning, it is a robot; If it owns biometric features and self-consciousness, it is still a robot, a so-called human-like robot. Only when it achieves an independent and intrinsic thinking space, could the robot be reckoned as a human being.

Table 1. When Robots could be regarded as Humans?

|  | Self-learning ability | Biometric Features | Self-consciousness | Independent and Intrinsic thinking space |
|---|---|---|---|---|
| Robots | √ | × | × | × |
| Robots | √ | √ | × | × |
| Robots | √ | √ | √ | × |
| Humans | √ | √ | √ | √ |

## More in-depth discussions

Instead of self-learning ability, biometric features and self-consciousness, the living space, particularly regarding the independent and intrinsic thinking space, is the most essential difference between humans and robots. If one day robots could own the independent and intrinsic thinking space like human beings, they could be regarded as humans. In turn, if human thinking space could be connected with cyberspace, and the prospect of Internet of Thinking is achieved by Human-Robot, humans could also be reckoned as robots [9]. By then, it may not be good news for humans.